\documentclass{article}

\usepackage{microtype}
\usepackage{graphicx}
\usepackage{booktabs} 

\usepackage{hyperref}

\usepackage[accepted]{icml2020}

\usepackage{macros}

\usepackage{cleveref}

\usepackage{subcaption}
\usepackage{enumitem}
\usepackage{relsize}
\usepackage{tcolorbox}

\icmltitlerunning{On Contrastive Learning for Likelihood-free Inference}

\begin{document}

\twocolumn[
\icmltitle{On Contrastive Learning for Likelihood-free Inference}




\begin{icmlauthorlist}
\icmlauthor{Conor Durkan}{ed}
\icmlauthor{Iain Murray}{ed}
\icmlauthor{George Papamakarios}{deep}
\end{icmlauthorlist}

\icmlaffiliation{ed}{School of Informatics, University of Edinburgh, United Kingdom}
\icmlaffiliation{deep}{DeepMind, London, United Kingdom}

\icmlcorrespondingauthor{Conor Durkan}{conor.durkan@ed.ac.uk}


\icmlkeywords{Machine Learning, ICML}

\vskip 0.3in
]



\printAffiliationsAndNotice{}  

\begin{abstract}

Likelihood-free methods perform parameter inference in stochastic simulator models where evaluating the likelihood is intractable but sampling synthetic data is possible. 
One class of methods for this likelihood-free problem uses a classifier to distinguish between pairs of parameter-observation samples generated using the simulator and pairs sampled from some reference distribution, which implicitly learns a density ratio proportional to the likelihood. 
Another popular class of methods fits a conditional distribution to the parameter posterior directly, and a particular recent variant allows for the use of flexible neural density estimators for this task. 
In this work, we show that both of these approaches can be unified under a general contrastive learning scheme, and clarify how they should be run and compared.

\end{abstract}

\section{Introduction} 
Modeling systems using parameterized stochastic simulators is prevalent across many scientific and engineering disciplines, including cosmology \citep{alsing-2019-fast}, high-energy physics \citep{brehmer-2018-constraining}, and computational neuroscience \citep{gonccalves-2019-training}. 
Specifying models in this way is appealing since it is often easier to describe a generative process implicitly rather than reasoning directly about an emergent probability distribution.
Traditional parameter inference algorithms such as variational methods and Markov Chain Monte Carlo (MCMC) usually don't apply to these models,
since explicit evaluation of the likelihood function is often intractable.

In recent years, significant progress has been made toward this challenge of likelihood-free inference \citep{sisson-2018-handbook}.
Most of the literature relies on sample-based methods that require very large numbers of simulations.
Methods based on neural networks have reduced the number of simulations needed to perform effective inference by orders of magnitude \citep{papamakarios-2019-phd, cranmer-2019-frontier}. Some of these can evaluate and sample from an approximate posterior directly, bypassing the need for MCMC schemes altogether.
Ultimately, this line of work seeks to develop a robust and practical toolkit which practitioners can rely upon for whichever use case they desire.

In this paper, we consider two recently proposed approaches to neural likelihood-free inference: the first uses classification to approximate density ratios proportional to the likelihood \citep{hermans-2019-sre}, whereas the second directly casts the problem as a conditional density estimation task \citep{greenberg-2019-apt}.
We demonstrate that these two methods, traditionally viewed as distinct and compared as such in the literature, are both instances of a more general contrastive learning scheme, and can thus be unified under a single framework.
Using this perspective, we directly compare the properties and behaviour of both algorithms, and offer practical recommendations for those wishing to use these methods.

\section{Background}
Given a vector of parameters $ \bftheta $, a stochastic simulator generates latent random numbers $ \bfz $, and produces observed data $ \bfx = \bfg(\bftheta, \bfz) $, where $ \bfg $ is some nonlinear function. 
The likelihood of the parameters $ \bftheta $ given observed data $ \bfx $ is then
\begin{align}
    p(\bfx \g \bftheta) = \int \delta(\bfx - \bfg(\bftheta, \bfz))\, p(\bfz \g \bftheta) \diff \bfz,
\end{align}
where $ \delta(\cdot) $ is the Dirac delta, and this integral is intractable in general. 
In some cases it may be possible to evaluate the joint likelihood $ p(\bfx, \bfz \g \bftheta) $, but (i) this requires performing inference on the constrained manifold $ \roundbr{\bfg(\bftheta, \bfz), \bfz} $, and (ii) often this augmented space is sufficiently high-dimensional to make standard inference approaches infeasible. 
In other cases, the simulator may be provided to us as a black-box whose internal workings are not accessible, and it is still desirable to carry out inference in this scenario. 

We are interested in computing the posterior $ p(\bftheta \g \bfx_{0}) $ for a specific observation $ \bfx_{0} $, where we have prior beliefs $ p(\bftheta) $ about the parameters.
This amounts to computing the density ratio $ \frac{p(\bfx_{0} \g \bftheta)}{p(\bfx_{0})} = \frac{p(\bftheta \g \bfx_{0})}{p(\bftheta)} $, and this quantity need only be known up to a constant independent of $ \bftheta $ if our approach uses MCMC schemes.
Throughout, we assume that the simulator defines a valid probability density function $ p(\bfx \g \bftheta) $ over observations $ \bfx $.

\subsection{Learning density ratios}
It is well known that binary classification can be used to learn density ratios \citep{hastie-2001-elements, sugiyama-2012-density-ratio, mohamed-2016-learning-in-implicit-generative-models}.
Defining $ p_{\text{pos}}(\bfx) = p(\bfx \g y \!=\! 1) $ and $ p_{\text{neg}}(\bfx) = p(\bfx \g y \!=\! 0) $ for binary $ y $, and assuming each class is equally likely a priori yields
\begin{align}
    p(y = 1 \g \bfx) = \frac{p_{\text{pos}}(\bfx)}{p_{\text{pos}}(\bfx) + p_{\text{neg}}(\bfx)} = \frac{r(\bfx)}{r(\bfx) + 1},
\end{align}
where $ r(\bfx) = \frac{p_{\text{pos}}(\bfx)}{p_{\text{neg}}(\bfx)} $. 
In other words, an optimal binary classifier which distinguishes between samples from the two distributions recovers the ratio between their density functions. 
A common choice of parameterization for this classifier is $ p(y = 1 \g \bfx) = \sigma(f_{\bfphi}(\bfx)) $, where $ \sigma(\cdot) $ is the logistic sigmoid and $ f_{\bfphi} $ is a real-valued function whose parameters $ \bfphi $ can be fit using maximum likelihood, and the optimal value for $ f_{\bfphi} $ is $ f_{\bfphi}(\bfx) = \log r(\bfx) $. 

In the context of likelihood-free inference, \citet{hermans-2019-sre} consider positive examples ($y=1$) those pairs $ (\bfx, \bftheta) \sim p(\bfx, \bftheta) $ which are distributed jointly, and generated by sampling a parameter from the prior and subsequently simulating an observation for that parameter setting. 
In contrast, negative examples  ($y=0$) are those pairs $ (\bfx, \bftheta) \sim p(\bfx)\,p(\bftheta) $ with parameters and observations sampled independently from their respective marginal distributions. 
Then binary classification recovers the ratio
\begin{align}
    r(\bfx, \bftheta) = \frac{p(\bfx, \bftheta)}{p(\bfx)\,p(\bftheta)} = \frac{p(\bfx \g \bftheta)}{p(\bfx)} = \frac{p(\bftheta \g \bfx)}{p(\bftheta)}.
\end{align}
Since the prior is known, it is possible to evaluate the parameter posterior exactly using $ p(\bftheta \g \bfx_{0}) = r(\bfx_{0}, \bftheta)\, p(\bftheta) $. 
However, only exact evaluation of the posterior distribution is achieved in this case, whereas sampling still requires an MCMC scheme. 
Likelihood-free inference by ratio estimation \citep[LFIRE,][]{thomas-2016-lfire} is a pioneering example of this approach.
However, \citet{thomas-2016-lfire} fit a classifier using only observations $ \bfx $ as input, meaning a separate classifier is required for each posterior evaluation $ p(\bftheta \vert \bfx) $.

Moreover, \citet{hermans-2019-sre} note that since we are interested in computing the posterior for a particular observation $ \bfx_{0} $, it may be wasteful to learn the ratio $ r(\bfx, \bftheta) $ for all pairs $ (\bfx, \bftheta) $.
Instead, it might be preferable to focus attention on those parameters which may have plausibly generated $ \bfx_{0} $, rather than parameters sampled from the prior. 
Suppose we have access to such a proposal prior, which we denote $ \tilde{p}(\bftheta) $. 
Then, considering positive examples $ (\bfx, \bftheta) \sim \tilde{p}(\bfx, \bftheta) = p(\bfx \g \bftheta)\, \tilde{p}(\bftheta) $ and negative examples $ (\bfx, \bftheta) \sim \tilde{p}(\bfx)\, \tilde{p}(\bftheta) $, binary classification recovers the ratio
\begin{align}
    \tilde{r}(\bfx, \bftheta) = \frac{\tilde{p}(\bfx, \bftheta)}{\tilde{p}(\bfx)\, \tilde{p}(\bftheta)} = \frac{p(\bfx \g \bftheta)}{\tilde{p}(\bfx)} 
\end{align}
At the expense of exact posterior evaluation, we have ostensibly gained a more accurate ratio estimate for the parameters corresponding to the observation of interest. 
Moreover, it is still possible to evaluate the posterior up to a constant, since $ p(\bftheta \g \bfx_{0}) \propto \tilde{r}(\bfx_{0}, \bftheta) \,p(\bftheta) $, so MCMC schemes can still be used to generate samples from the desired posterior. 
Using this approach, an algorithm emerges where the likelihood ratio is refined iteratively across a series of rounds, and training data for a given round is generated using the estimated ratios from preceding rounds i.e.\ the proposal $ \tilde{p}(\bftheta) $ is a mixture of posterior estimates from all preceding rounds. 
We use the term Sequential Ratio Estimation (SRE) to describe this approach of \citet{hermans-2019-sre}.

Finally, binary classification can equivalently be viewed as a task where the model must correctly identify the correct example given a set consisting of exactly one positive example drawn jointly with an observation, and one negative example drawn marginally.
This equivalent perspective will prove useful in deriving a unifying notation in later sections. 
Denoting this set $ \Theta = \curlybr{\bftheta_{0}, \bftheta_{1}} $ and assuming each class is equally likely a priori, we have
\begin{align}
    \!\!\!p(y = 1 \g \bfx, \Theta) &= \frac{p(\bfx, \Theta \g y = 1)}{p(\bfx, \Theta \g y = 0) + p(\bfx, \Theta \g y = 1)} \\
    &= \frac{p(\bftheta_{1} \g \bfx)\,p(\bftheta_{0})}{p(\bftheta_{0} \g \bfx)\,p(\bftheta_{1}) + p(\bftheta_{1} \g \bfx)\,p(\bftheta_{0})} \\
    &= \frac{p(\bftheta_{1} \g \bfx) / p(\bftheta_{1})}{p(\bftheta_{0} \g \bfx) / p(\bftheta_{0}) + p(\bftheta_{1} \g \bfx) / p(\bftheta_{1})}. \label{eqn:binary-ratio}
\end{align}
Now if we parameterize a binary classifier by
\begin{align}
    \!\!\!p(y = 1 \g \bfx, \Theta) = \frac{\exp(f_{\bfphi}(\bftheta_{1}, \bfx))}{\exp(f_{\bfphi}(\bftheta_{0}, \bfx)) + \exp(f_{\bfphi}(\bftheta_{1}, \bfx))}, \label{eqn:parameterized-binary-ratio}
\end{align}
then the optimal classifier has $ f_{\bfphi}(\bftheta, \bfx) = \log \frac{p(\bftheta \g \bfx)}{p(\bftheta)} + c(\bfx) $, where $ c(\bfx) $ is constant with respect to $ \bftheta $. 

\subsection{Learning the posterior}
Likelihood-free inference can also be cast as a conditional density estimation problem directly. Consider a parameterized conditional density estimator $ q_{\bfphi}(\bftheta \g \bfx) $, sometimes referred to as a recognition model. The parameters $ \bfphi $ can be fit by minimizing
$ \mathcal{L}(\bfphi) = \mathbb{E}_{(\bfx, \bftheta) \sim p(\bfx, \bftheta)}\squarebr{- \log q_{\bfphi}(\bftheta \g \bfx)} $. 
With enough data and a sufficiently flexible model, $ q_{\bfphi}(\bftheta \g \bfx) $ will converge to the posterior density $ p(\bftheta \g \bfx) $ for all pairs $ (\bfx, \bftheta) $.
Then the posterior density of interest can be computed using $ p(\bftheta \g \bfx_{0}) = q_{\bfphi}(\bftheta \g \bfx_{0}) $.
However, since we are again only interested in the posterior for a specific observation $ \bfx_{0} $, it may be beneficial to learn the conditional density more carefully for parameters which could have plausibly generated that observation.
Indeed, recent work, which we group under the umbrella of Sequential Neural Posterior Estimation \citep[SNPE,][]{papamakarios-2016-snpe-a, lueckmann-2017-snpe-b, greenberg-2019-apt}, has developed sequential procedures for exactly this purpose, narrowing focus to parameters of interest across a series of rounds, where the posterior estimate for a given round is used as the proposal prior for the subsequent round.
Central to the function of these methods is the proposal posterior $ \tilde{p}(\bftheta \g \bfx) $, related to the true posterior by
\begin{align}
    \tilde{p}(\bftheta \g \bfx) = p(\bftheta \g \bfx) \frac{\tilde{p}(\bftheta)}{p(\bftheta)} \frac{p(\bfx)}{\tilde{p}(\bfx)}.
\end{align}

\textbf{SNPE-A\@.} The first method is known as SNPE-A \citep{papamakarios-2016-snpe-a}. 
Given a proposal prior $ \tilde{p}(\bftheta) $, SNPE-A fits a neural density estimator $ q_{\bfphi}(\bftheta \g \bfx) $ by minimizing $ \mathcal{L}(\bftheta) = \mathbb{E}_{(\bfx, \bftheta) \sim p(\bfx \g \bftheta) \tilde{p}(\bftheta)}\squarebr{- \log q_{\bfphi}(\bftheta \g \bfx)} $, converging to the proposal posterior $ \tilde{p}(\bftheta \g \bfx) $. 
The relation $ p(\bftheta \g \bfx) \propto \tilde{p}(\bftheta \g \bfx) \frac{p(\bftheta)}{\tilde{p}(\bftheta)} = q_{\bfphi}(\bftheta \g \bfx) \frac{p(\bftheta)}{\tilde{p}(\bftheta)} $ can then be used to compute the desired conditional density up to a constant. 
In certain cases, we can compute the posterior analytically.
For example, when both  $ p(\bftheta) $ and $ \tilde{p}(\bftheta) $ are Gaussian, and the density estimator $ q_{\bfphi}(\bftheta \g \bfx) $ is a mixture density network \citep{bishop-1994-mdn} with $ M $ Gaussian components, the posterior is also a mixture of Gaussians with $ M $ components. 
However, the algorithm precludes the use of more flexible conditional density estimators, such as autoregressive models \citep{uria-2016-nade} or normalizing flows \citep{papamakarios-2019-flows-review}, if we want a tractable posterior. 

\textbf{SNPE-B\@.} The second of these methods, referred to as SNPE-B \citep{lueckmann-2017-snpe-b}, allows for the use of arbitrary neural density estimators.
This is achieved through minimizing an importance-weighted loss $ \mathcal{L}(\bfphi) =  \mathbb{E}_{(\bfx, \bftheta) \sim p(\bfx \g \bftheta) \tilde{p}(\bftheta)}\squarebr{- \frac{p(\bftheta)}{\tilde{p}(\bftheta)} \log q_{\bfphi}(\bftheta \g \bfx)} $, where $ q_{\bfphi}(\bftheta \g \bfx) $ now recovers the correct posterior directly.
Unfortunately, the importance weights $ \frac{p(\bftheta)}{\tilde{p}(\bftheta)} $ are often high variance, and can lead to poor performance of the algorithm overall.  

\textbf{SNPE-C (APT)\@.} Finally, the third method, and of particular interest in this paper, is Automatic Posterior Transformation \citep[APT,][]{greenberg-2019-apt}, or SNPE-C in the existing taxonomy. 
SNPE-C neatly reparameterizes the proposal posterior objective from SNPE-A so that maximum likelihood recovers the true posterior directly.  
Setting 
\begin{align}
    \tilde{q}_{\bfphi}(\bftheta \g \bfx) = q_{\bfphi}(\bftheta \g \bfx) \frac{\tilde{p}(\bftheta)}{p(\bftheta)} \frac{1}{Z_{\bfphi}(\bfx)},
\end{align}
where $ Z_{\bfphi}(\bfx) = \int q_{\bfphi}(\bftheta \g \bfx) \frac{\tilde{p}(\bftheta)}{p(\bftheta)} \diff \bftheta $, SNPE-C minimizes $ \mathcal{L}(\bfphi) = \mathbb{E}_{(\bfx, \bftheta) \sim p(\bfx \g \bftheta) \tilde{p}(\bftheta)}\squarebr{- \log \tilde{q}_{\bfphi}(\bftheta \g \bfx)} $. 

Unfortunately, SNPE-C as described requires the ability to calculate the normalizing constant $ Z_{\bfphi}(\bfx) $, which can be computed analytically for an MDN, but not in general for more flexible density estimators such as normalizing flows. 
However, \citet{greenberg-2019-apt} note that if the proposal is uniform and supported on a finite set of discrete `atoms' $ \Theta $, the integral reduces to a sum, and the proposal posterior now defines a distribution over this discrete set:
\begin{align}
    \tilde{q}_{\bfphi}(\bftheta \g \bfx) = \frac{q_{\bfphi}(\bftheta \g \bfx) / p(\bftheta)}{\sum_{\bftheta^{\prime} \in \Theta} q_{\bfphi}(\bftheta^{\prime} \g \bfx) / p(\bftheta^{\prime})}. \label{eqn:apt-ratio}
\end{align}
Moreover, \citet{greenberg-2019-apt} show that when the contrasting set is drawn from a distribution over parameters whose support contains the support of the true posterior, the density estimator $ q_{\bfphi}(\bftheta \g \bfx) $ still converges to the full continuous posterior.  
Intuitively, the inference problem has been rephrased as a series of multiple choice questions: given an observation $ \bfx $ and a set of possible parameters $ \Theta $ which may have generated that observation, the task is to identify the correct parameter $ \bftheta $. 
For the rest of the paper,
SNPE-C refers specifically to this atomic variant of the algorithm.
\looseness=-1

\section{Method}

\begin{algorithm*}[tb]
   \caption{(Sequential) Contrastive Likelihood-free Inference}
   \label{alg:contrastive-lfi}
\begin{algorithmic}[1]
   \STATE {\bfseries Input:} Simulator $ p(\bfx \g \bftheta) $, prior $ p(\bftheta) $, true observation $\bfx_{0} $, classifier $ f_{\bfphi}(\bftheta, \bfx) $, rounds $ R $, simulations per round $ N $.
  \COMMENT{For SNPE-C, we require $ f_{\bfphi}(\bftheta, \bfx) = \log \frac{q_{\bfphi}(\bftheta \g \bfx)}{p(\bftheta)} $, where $ q_{\bfphi}(\bftheta \g \bfx) $ is a normalized conditional density estimator.}
   \STATE {\bfseries Initialize:} Posterior $ p^{(1)}(\bftheta) = p(\bftheta) $, dataset $ \mathcal{D} = \curlybr{} $.
   \FOR{$r=1$ {\bfseries to} $R$}
   \STATE Sample $ \bftheta^{(n)} \sim p^{(r)}(\bftheta) $, $ n = 1 \dots N $.
  \COMMENT{For SRE, this step requires MCMC, but in the case of SNPE-C we can sample i.i.d.\ from the density estimator, using rejection sampling if the density estimator has support outside the prior.}
   \STATE Simulate $ \bfx^{(n)} \sim p(\bfx \g \bftheta^{(n)}) $, $ n = 1 \dots N $.
   \STATE Aggregate training data $ \mathcal{D} = \mathcal{D} \cup \curlybr{ \roundbr{ \bfx^{(n)}, \bftheta^{(n)} } }_{n = 1}^{N} $.
   \WHILE{$ f_{\bfphi} $ not converged}
   \STATE Sample mini-batch $ \curlybr{ \roundbr{ \bfx^{(b)}, \bftheta^{(b)} } }_{b = 1}^{B} \sim \mathcal{D} $. 
  \COMMENT{This is equivalent to sampling parameters from a proposal $ \bftheta \sim \tilde{p}(\bftheta) = \frac{1}{r} \sum_{r^{\prime} = 1}^{r} p^{(r^{\prime})}(\bftheta) $, and then simulating $ \bfx \sim p(\bfx \g \bftheta) $.}
   \STATE For each pair in the mini-batch, sample $ 0 < K - 1 < B $ contrasting parameters without replacement from the remainder of the mini-batch.  
  \COMMENT{Equivalent to sampling each contrasting parameter independently from $ \tilde{p}(\bftheta) $.}
   \STATE Update $ \bfphi $ by stochastic gradient descent using objective $ \mathcal{L}(\bfphi) = - \frac{1}{B} \sum_{b = 1}^{B} \log \frac{\exp \roundbr{ f_{\bfphi}(\bftheta^{(b)}, \bfx^{(b)}) }}{\sum_{k=1}^{K} \exp \roundbr{ f_{\bfphi}(\bftheta^{(k)}, \bfx^{(b)}) }} $.
   \ENDWHILE
  \STATE Update posterior $ p^{(r)}(\bftheta) \propto \exp(f_{\bfphi}(\bftheta, \bfx_{0})) $. 
  \COMMENT{For SNPE-C, we can set $ p^{(r)}(\bftheta) = q_{\bfphi} (\bftheta \g \bfx_{0}) $ directly.}
   \ENDFOR
\end{algorithmic}
\end{algorithm*}

\textbf{Unifying SRE and SNPE-C\@.} 
There is clear similarity between \cref{eqn:parameterized-binary-ratio} and \cref{eqn:apt-ratio}; indeed, modulo the switch from a binary to multi-class setting, the difference is only notational.
SRE and SNPE-C are in fact both instances of a more general contrastive learning scheme.
To see this, consider the posterior over labels for $ K $ categories given an observation $ \bfx $ and a set of examples $ \Theta $ containing exactly one positive example $ \bftheta $ which generated that observation, while the remaining set of contrasting examples $ \Theta^{(\setminus k)} $ are drawn jointly from some other distribution conditioned on this joint pair. Then, again assuming each class is equally likely a priori, we have 
\begin{align}
    p(y = k \g \bfx, \Theta) &= \frac{p(\bfx, \Theta \g y = k)\,p(y = k)}{\sum_{k^{\prime}} p(\bfx, \Theta \g y = k^{\prime})\,p(y = k^{\prime})} \\
    &= \frac{p(\bftheta^{(k)} \g \bfx)\,p(\Theta^{(\setminus k)} \g \bfx, \bftheta^{(k)})}{\sum_{k^{\prime}} p(\bftheta^{(k^{\prime})} \g \bfx)\,p(\Theta^{(\setminus k^{\prime})} \g \bfx, \bftheta^{(k^{\prime})})}.\notag
\end{align}
Now suppose the elements of the contrasting set are drawn independently of both the joint pair $ \roundbr{\bfx, \bftheta^{(k)}} $, and of each other, by sampling $ K - 1 $ times from the prior. In this case, $ p(\Theta^{(\setminus k)} \g \bfx, \bftheta^{(k)}) = \prod_{l \neq k} p(\bftheta^{(l)}) $, and we have
\begin{align}
    p(y = k \g \bfx, \Theta)
    &= \frac{p(\bftheta^{(k)} \g \bfx) \prod_{l \neq k} p(\bftheta^{(l)})}{\sum_{k^{\prime}} p(\bftheta^{(k^{\prime})} \g \bfx) \prod_{l \neq k^{\prime}} p(\bftheta^{(l)})} \\
    &= \frac{p(\bftheta^{(k)} \g \bfx) / p(\bftheta^{(k)})}{\sum_{k^{\prime}} p(\bftheta^{(k^{\prime})} \g \bfx) / p(\bftheta^{(k^{\prime})})} \label{eqn:14} \\ 
    &= \frac{\exp\roundbr{f_{\bfphi}(\bftheta^{(k)}, \bfx) + c(\bfx)}}{\sum_{k^{\prime}} \exp\roundbr{f_{\bfphi}(\bftheta^{(k^{\prime})}, \bfx) + c(\bfx)}}, \label{eq:contrastive-learning-scheme}
\end{align}
where the progression from \cref{eqn:14} to \cref{eq:contrastive-learning-scheme} is directly analogous to the progression from \cref{eqn:binary-ratio} to \cref{eqn:parameterized-binary-ratio}.
Once more, we see that the optimal classifier implicitly learns the desired log-density ratio up to proportionality. 
Moreover, we still recover the correct ratio even if we sample parameters from a proposal $ \tilde{p}(\bftheta) $ instead of the prior $ p(\bftheta) $, since $ \frac{\tilde{p}(\bftheta \g \bfx)}{\tilde{p}(\bftheta)} \propto \frac{p(\bftheta \g \bfx)}{p(\bftheta)} $.
With this result, the connection between SRE and SNPE-C becomes clear.
\begin{itemize}[leftmargin=*, noitemsep, topsep=0pt]
\itemsep1em 
    \item Setting $ K \!=\! 2 $ and parameterizing the classifier as a standard feed-forward neural network $ f_{\bfphi}(\bftheta, \bfx) = \text{NN}_{\bfphi}(\bftheta, \bfx) $ recovers SRE\@. The derivation above also generalizes SRE from the binary case to multiple classes, and we will show this generally improves performance of the algorithm. 
    \item The multi-class problem with parameterization of the classifier in terms of a density estimator $ f_{\bfphi}(\bftheta, \bfx) = \log \frac{q_{\bfphi}(\bftheta \g \bfx)}{p(\bftheta)} $ recovers SNPE-C\@. In this case, $ q_{\bfphi}(\bftheta \g \bfx) $ yields a normalized approximation to the posterior when the density and the prior are supported on the same set (see \cref{correctness-of-apt} for details), and may allow for efficient and exact evaluation and sampling depending on the particular choice of density estimator. This provides an alternative derivation to that given by \citet{greenberg-2019-apt} of the correctness of the atomic variant of SNPE-C\@.  
\end{itemize}{}
We believe this unifying perspective is beneficial to both practitioners and researchers, since comparison of these methods can now be cast in a more direct way, and the trade-offs of each more carefully considered.  
For instance, immediate questions arise concerning choice of contrasting set size, and whether reparameterizing the classifier in terms of a density estimator impacts inference.
It is also interesting to note how two lines of research into the likelihood-free problem converged essentially independently to very similar ideas; indeed, \citet{hermans-2019-sre} compare to SNPE-C in their experiments without explicitly noting this connection.
We present the general algorithm which encompasses both of these methods, iteratively refining a posterior approximation across a sequence of rounds, in \cref{alg:contrastive-lfi}.

\textbf{Connection to mutual information estimation.}
Similar observations concerning density ratios have previously been made in the wider machine learning literature, but the implications for likelihood-free inference may not have been fully appreciated.
The topic has received particular attention recently due to interest in mutual information estimation with neural networks \citep{belghazi-2018-mine, oord-2018-cpc, poole-2019-variational-bounds}. 
In this context, the classifier is often referred to as a critic, and multi-sample variational bounds have been derived for the mutual information, where it is also known that it can be advantageous to reparameterize the critic in terms of known densities. 
With this work, we seek to clarify these points within the likelihood-free inference community, and unify two promising lines of research which until now have been presented as distinct.

Of particular note is the connection of this contrastive learning scheme for likelihood-free inference to the multi-sample lower bound on the mutual information popularized by \citet{oord-2018-cpc}. 
Imagine we sample a mini-batch of $ B $ pairs $ \curlybr{(\bfx^{(b)}, \bftheta^{(b)})}_{b = 1}^{B} $ from our training set, where $ \bftheta^{(b)} \sim \tilde{p}(\bftheta) $ was sampled from the some proposal, and $ \bfx^{(b)} \sim p(\bfx \g \bftheta^{(b)}) $ was generated using the simulator. 
To generate a contrasting set of examples, we choose the remaining parameters in the batch, in effect sampling marginally from the proposal $ \tilde{p}(\bftheta) $.
Indeed, this is the procedure used by SNPE-C to generate a contrasting set, and SRE can be extended similarly.
Equivalently, this amounts to setting $ K \!=\! B $ in \cref{eq:contrastive-learning-scheme}.
The classification loss for a minibatch is then  
\begin{align}
    \mathcal{L}_{B}(\bfphi) = -\frac{1}{B} \sum_{b = 1}^{B} \log \frac{\exp\roundbr{f_{\bfphi}(\bftheta^{(b)}, \bfx^{(b)})}}{\sum_{b^{\prime}} \exp\roundbr{f_{\bfphi}(\bftheta^{(b^{\prime})}, \bfx^{(b)})}}.
\end{align}
The negative of this quantity is a well-known multi-sample lower bound on the mutual information between the two distributions defined by the joint $ \tilde{p}(\bfx, \bftheta) = p(\bfx \g \bftheta)\,\tilde{p}(\bftheta) $ and the product of marginals $ \tilde{p}(\bfx)\, \tilde{p}(\bftheta) $ \citep{poole-2019-variational-bounds}, so that optimization of the classification loss can be equivalently interpreted as optimization of an estimate of the mutual information between these distributions.

This perspective gives some motivation for a sequential approach in likelihood-free methods, where in each round we sample parameters from a successively more accurate targeted proposal $ \tilde{p}(\bftheta) $ rather than the prior, which may be expected to decrease the mutual information across rounds. 
It also suggests that increasing the size of the contrasting set $ K $ may be beneficial, since this mutual information estimate is bounded above by $ \log K $.
However, neither SRE or SNPE-C needs to calculate the mutual information explicitly, and the density ratios can be learned for any $ K \geq 2 $.

\begin{figure*}[t]
	\centering
    \begin{subfigure}[b]{0.5\textwidth}
        \centering
        \includegraphics[width=0.9\textwidth]{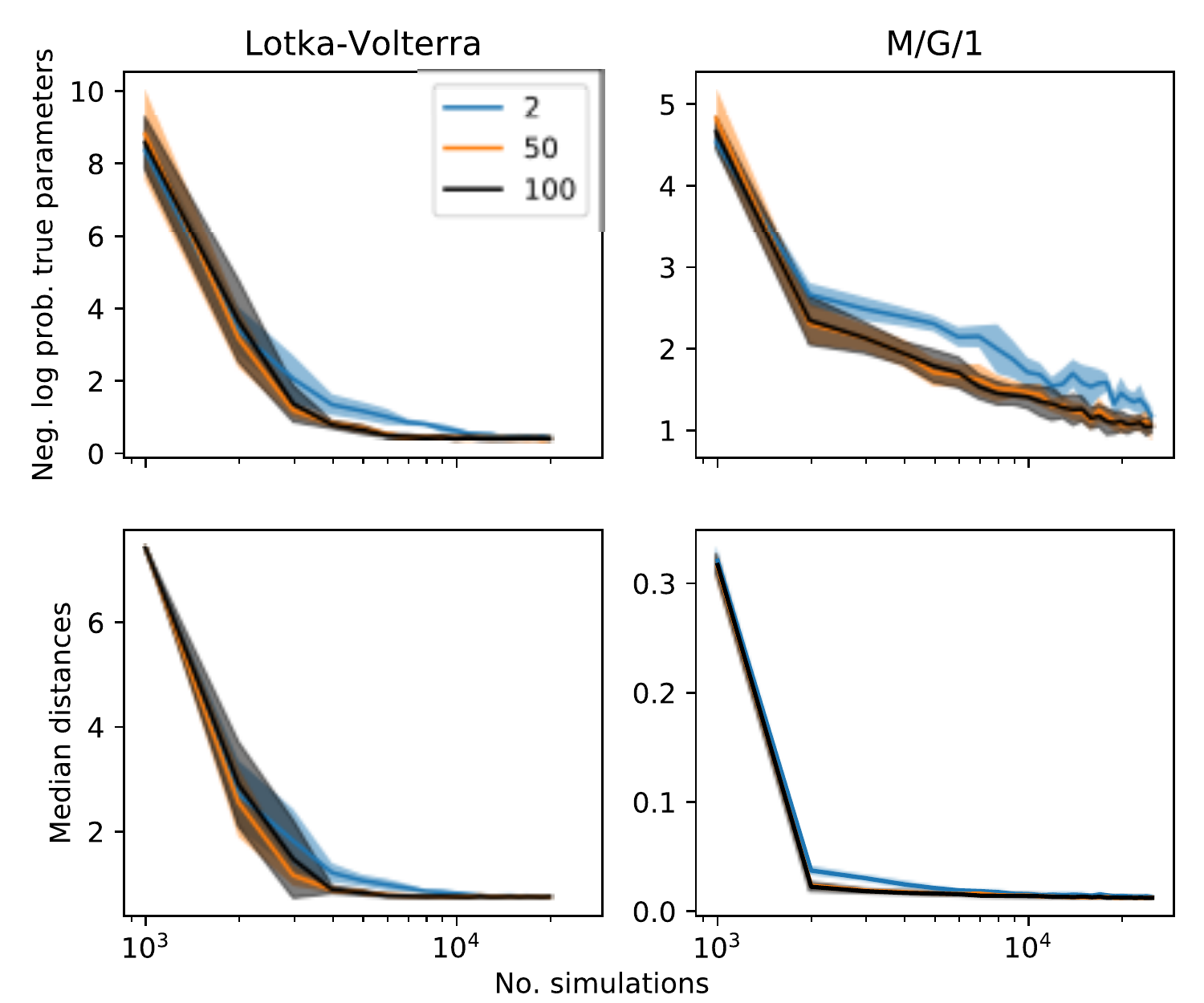}
    \end{subfigure}%
    \begin{subfigure}[b]{0.5\textwidth}
        \centering
        \includegraphics[width=0.9\textwidth]{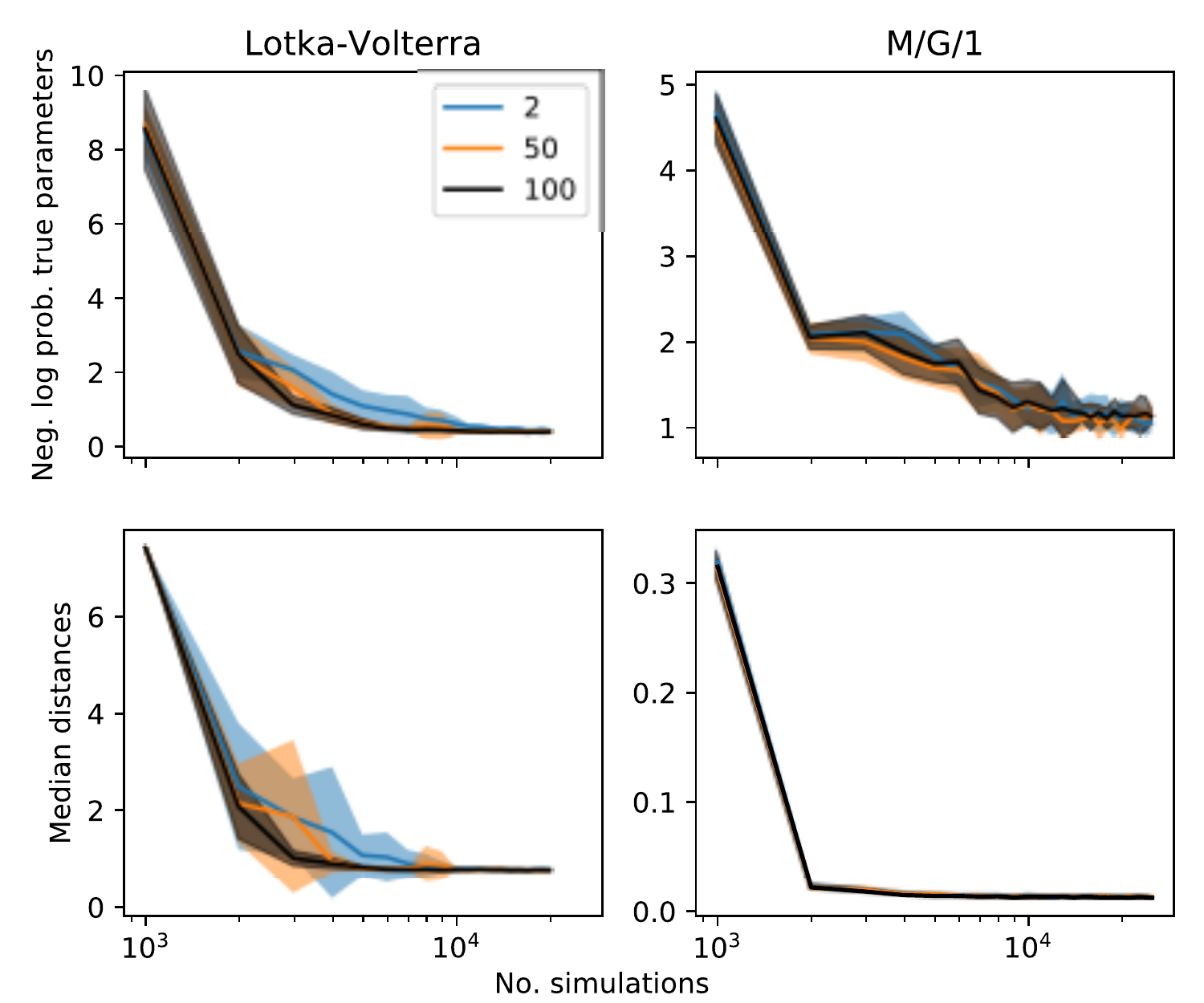}
    \end{subfigure}
    \vspace{-10pt}
	\caption{Comparison of SRE (left) and SNPE-C (right) across tasks and metrics, featuring mean and standard deviation across 10 random seeds. Negative log probability of true parameters is calculated using a kernel density estimate on posterior samples at the end of each round. Median distance uses the L2 distance between observations generated at each round and true observations. Increasing the contrasting-set size leads to more efficient inference, if only marginally in some cases.}
	\label{fig:size-of-contrasting-set}
	\vspace{-10pt}
\end{figure*}

\section{Related Work}

\textbf{Approximate Bayesian Computation (ABC)\@.} 
We use the term ABC to describe the group of likelihood-free inference algorithms which require (i) choice of summary statistics $ \overline{\bfx} $ for observations $ \bfx $, (ii) choice of a metric $ d(\cdot, \cdot) $ which computes distances between summary statistics, and (iii) choice of a tolerance $ \varepsilon > 0 $ which determines whether two summary statistics are sufficiently similar based on the computed distance. 
Broadly, ABC methods draw approximate posterior samples by first proposing according to some scheme, and then rejecting samples whose summary statistics are distance more than $\epsilon$ from the observed summaries.
Such algorithms have formed a central component of likelihood-free inference research for many years;  see e.g.\ reviews by \citet{beaumont-2010-abc_evo_eco}, \citet{lintusaari-2016-abcreview} and \citet{beaumont-2019-abcreview}. 
A drawback of ABC methods is that they generally require a large number of simulations to yield accurate results, which can be prohibitive when the simulator is expensive.
Comparisons in the existing literature suggest that likelihood-free methods based on neural density estimators can outperform traditional approaches such as Sequential Monte Carlo ABC \citep{sisson-2007-smc_abc, beaumont-2009-smc_abc, toni-2009-smc_abc} in terms of simulation cost, often requiring orders of magnitude fewer simulations \citep{papamakarios-2016-snpe-a, papamakarios-2018-snl}.

\textbf{Learning the likelihood.} 
Conversely to fitting a conditional density estimator to the posterior directly, it is also possible to fit a synthetic likelihood to approximate $ p(\bfx \vert \bftheta) $. A synthetic likelihood can be estimated separately for each parameter $\bftheta$ considered, e.g.\ as part of an MCMC scheme \citep{wood-2010-sl, price-2018-bayesian_sl, fasiolo-2018-saddlepoint}, or it can be amortized across parameters \citep{fan-2013-abc, meeds-2014-gps}.
One advantage of the likelihood-learning approach is that parameters can be drawn from any proposal
\citep{papamakarios-2018-snl}, and may even be chosen using active learning \citep{gutmann-2016-bolfi, lueckmann-2018-emulator-networks, jarvenpaa-2018-efficient}.
In addition, the simulator may be such that learning the likelihood is easier than learning the posterior directly, which is often the case when the forward process is specified using simple probabilistic primitives, as in probabilistic programming \citep{gordon-2014-probprog}. Finally, a model of the likelihood is often easier to diagnose than a model of the posterior, since exact samples from $ p(\bfx \vert \bftheta) $ can be readily obtained \citep{dalmasso-2019-validation}.
On the other hand, any methods based on a surrogate likelihood require an additional inference step (such as MCMC) for posterior sampling, which adds complexity to the overall algorithm and can introduce further approximation error.
Compounding this drawback is the fact that observations $ \bfx $ are often of much higher dimension than parameters $ \bftheta $, making conditional density estimation of observations a more difficult task.
Indeed, \citet{greenberg-2019-apt} show that the performance of the Sequential Neural Likelihood algorithm described by \citet{papamakarios-2018-snl} degrades when concatenating uninformative noise to observations $ \bfx $.

\textbf{Contrastive learning.} 
The use of classification for density-ratio estimation is long-standing in machine learning, and is closely linked to the broad paradigm of learning-by-comparison, or contrastive learning.
A prominent example of this line of thinking is Noise Contrastive Estimation \citep{gutmann-2010-nce}, a method for fitting unnormalized distributions, or energy-based models, by learning relative probabilities through comparison with a known noise distribution. 
This approach forms the basis of the subsequent LFIRE method \citep{thomas-2016-lfire}.
The branch of Generative Adversarial Networks literature concerned with fitting discriminators which approximate $ f $-divergences \cite{goodfellow-2014-gans, nowozin-2016-fgan} has also made significant contribution to the ubiquity of density-ratio estimation techniques.
Recently, contrastive learning has also shown great promise in unsupervised pre-training for vision tasks.
A variant of Contrastive Predictive Coding \citep{oord-2018-cpc, henaff-2019-cpc-extended} has demonstrated that a linear classifier on top of features learned in an unsupervised manner on ImageNet outperforms early convolutional networks in accuracy, while Momentum Contrast \citep{he-2019-momentum-contrast} outperforms supervised pre-training in a series of downstream detection and segmentation tasks. 

\textbf{Density-ratio estimation for likelihood-free inference.} 
The ratio of the likelihood $ p(\bfx \vert \bftheta) $ to the marginal $ p(\bfx) $ is of central importance to Bayesian inference, and thus a natural quantity to estimate in likelihood-free inference. 
An early example of estimating this ratio is given by \mbox{\citet{izbicki-2014-density-ratio}}; subsequent examples include LFIRE \citep{thomas-2016-lfire} and SRE \citep{hermans-2019-sre}, which we have already discussed. 
Another possibility is to learn the ratio between likelihoods $ p(\bfx \vert \bftheta_1) $ and $ p(\bfx \vert \bftheta_2) $, with or without amortization across parameters $\bftheta_1$ and/or $\bftheta_2$. 
This likelihood ratio can be used as the acceptance ratio in MCMC \citep{pham-2014-abcmcmc} or as a test statistic in hypothesis testing \citep{brehmer-2018-constraining}, and can be learned by binary classification \citep{cranmer-2016-carl} or by regressing the joint-likelihood ratio if the latter is available \citep{brehmer-2018-mining, stoye-2018-cross-entropy}. 
Other applications of density-ratio estimation in likelihood-free inference include: using binary-classification performance as a distance metric for ABC \citep{gutmann-2018-classification}, estimating variational bounds \citep{tran-2017-hierarchical}, and testing goodness-of-fit of synthetic likelihoods \citep{dalmasso-2019-validation}.

\section{Experiments}
In this section we compare SRE and SNPE-C experimentally, and discuss practical details of their implementation.
For completeness we also consider Sequential Neural Likelihood \citep[SNL,][]{papamakarios-2018-snl}, another sequential method for likelihood-free inference which fits a neural density estimator as a surrogate likelihood.
The classifier for SRE is a feed-forward residual network with two residual blocks of 50 hidden units. 
All density estimators for SNL and SNPE-C use a MAF \citep{papamakarios-2017-maf} architecture, consisting of a stack of five MADEs \citep{germain-2015-made}, each with two hidden layers of 50 units, and a standard normal base distribution. 
While MAF features density evaluation that is efficiently parallelizable across data dimensions, sampling is sequential, but the dimensionality of the parameter spaces considered is low enough that sampling is efficient.
It is also worthwhile to point out the modularity of both SNL and SNPE-C, which readily allow for any off-the-shelf density estimator to be plugged in; the choice of MAF was made here in line with previous work.

We use axis-aligned slice sampling \citep{neal-2003-slice} as an MCMC method for SNL and SRE\@.
A single MCMC chain persists across rounds for each method, where we perform burn-in of 200 iterations whenever the target distribution changes, and retain every tenth accepted sample. 
In each round, the parameters of each method are fit using stochastic gradient descent with the Adam \citep{kingma-2014-adam} optimizer, a learning rate of 5e-4, and a mini-batch size of 100.
To prevent overfitting, we perform early-stopping based on a held-out validation set of ten percent of the training data aggregated so far, stopping training when validation performance does not improve over 20 epochs. 
We compare likelihood-free methods using a testbed of three simulators, namely the Nonlinear Gaussian simulator with tractable likelihood described by \citet{papamakarios-2018-snl}, along with the Lotka--Volterra predator-prey and M/G/1 queue models whose setups are detailed by \citet{papamakarios-2016-snpe-a}.
For all tasks we acquire 1000 new simulations per round, running the Nonlinear Gaussian and M/G/1 tasks for 25 rounds, and Lotka--Volterra for 20.
All experiments are repeated across 10 random seeds using a single GPU, and code is available at \url{https://github.com/conormdurkan/lfi}.

\begin{figure}[t]
	\centering
    \includegraphics[width=0.475\textwidth]{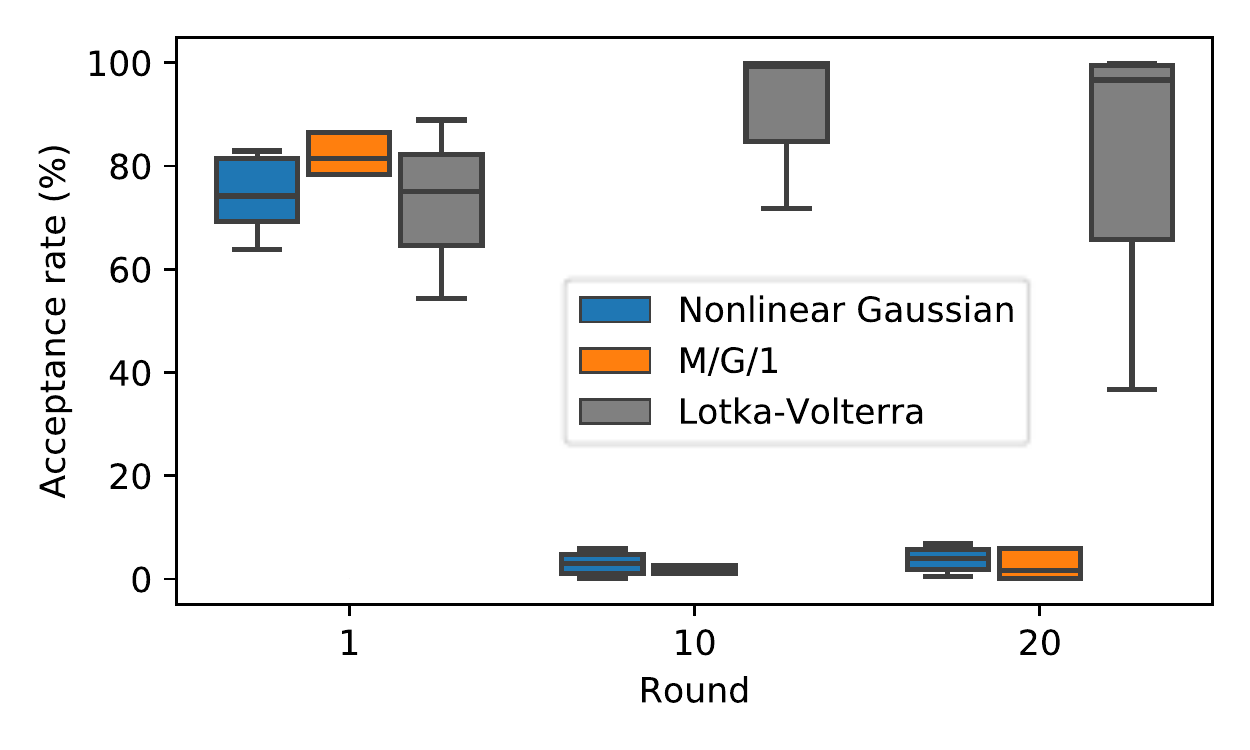}
    \vspace{-10pt}
	\caption{Distribution of rejection sampling acceptance rates for SNPE-C using $10^{7}$ samples in rounds 1, 10, and 20 across 10 random seeds. Whiskers delineate $ 5^{th} $ and $ 95^{th} $ percentiles. Maximum likelihood training in the first round forces $ q_{\bfphi} $ to place density in the prior support. However, learning by classification in later rounds means there is no penalty for leaking significant density, and this phenomenon occurs in two of three tasks, with high variance in the third.}
	\label{fig:acceptance-rates-snpe-c}
	\vspace{-15pt}
\end{figure}

\begin{figure*}[t]
    \vspace{-4pt}
	\centering
    \begin{subfigure}[b]{0.5\textwidth}
        \centering
        \includegraphics[width=0.9\textwidth]{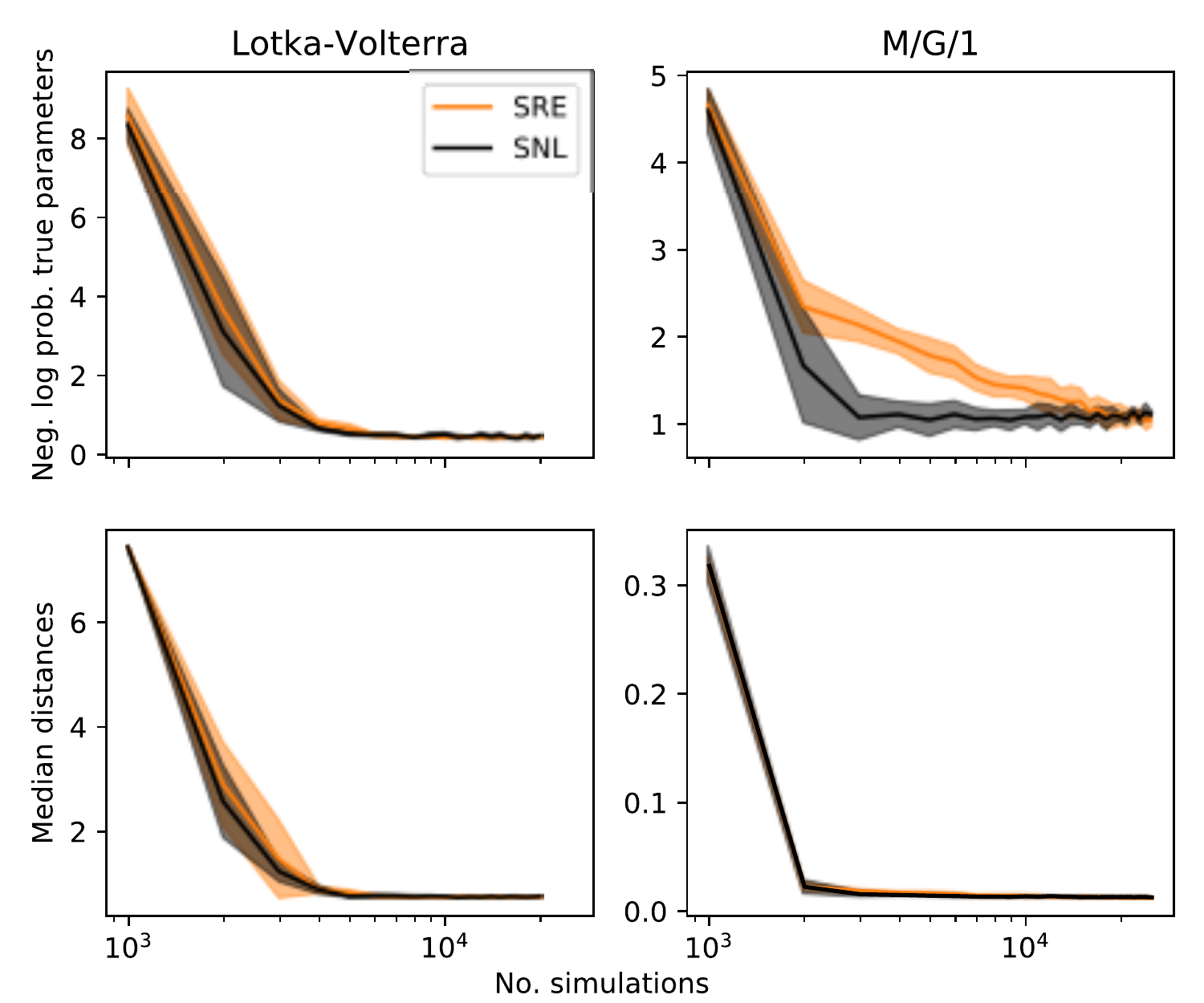}
    \end{subfigure}%
    \begin{subfigure}[b]{0.5\textwidth}
        \centering
        \includegraphics[width=0.9\textwidth]{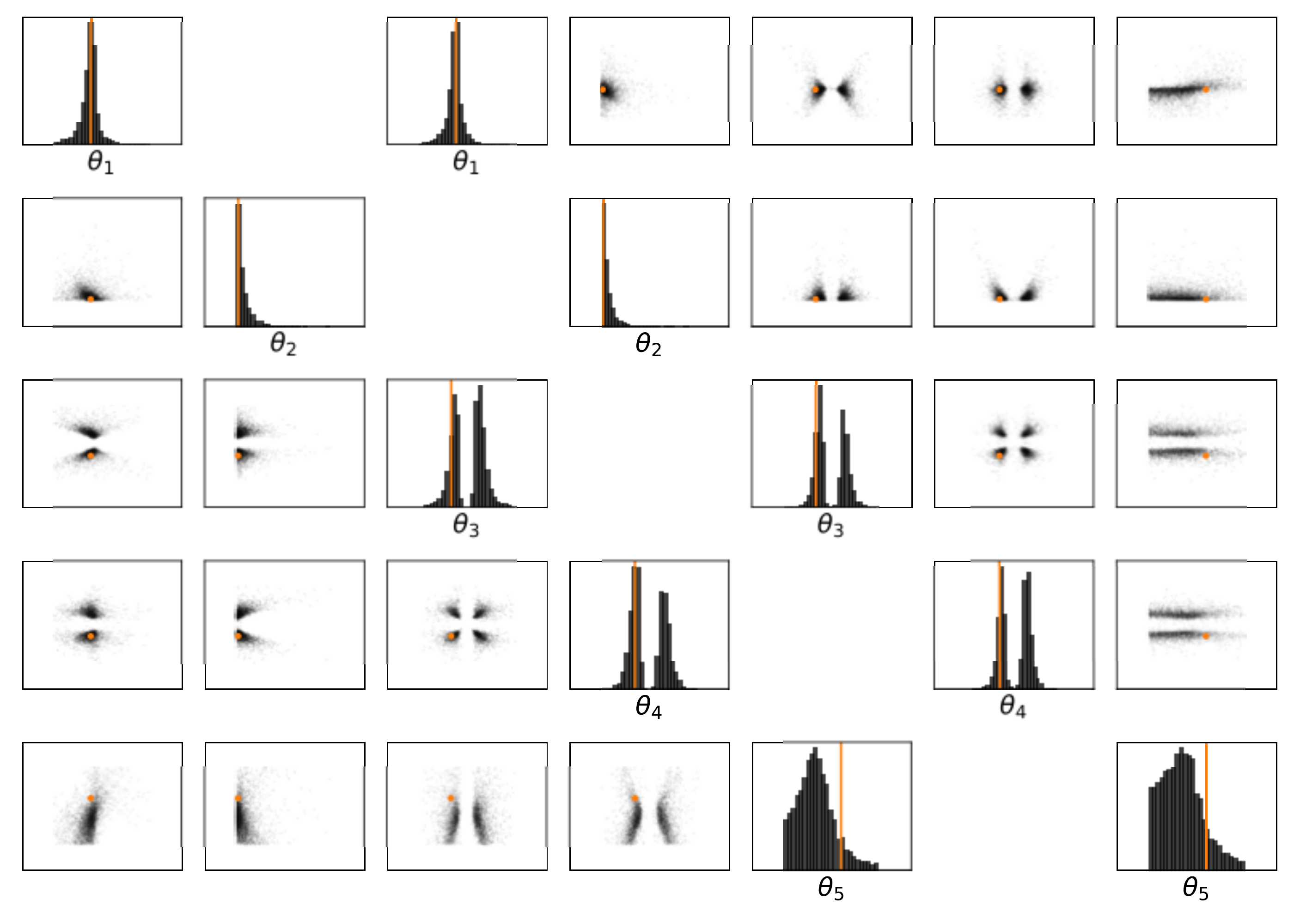}
    \end{subfigure}
    \vspace{-15pt}
	\caption{\textbf{Left}: Comparison of SRE and SNL across tasks and metrics, with negative log probability and median distance as in \cref{fig:size-of-contrasting-set}. \textbf{Right}: Posterior samples for SNL (left) and SRE (right) on the Nonlinear Gaussian task; true parameters in orange.}
	\label{fig:snl-sre-comparison}
	\vspace{-6pt}
\end{figure*}

\textbf{A note on choice of prior.}
A subtle detail arises when using SNPE-C with a density estimator whose support does not match that of the prior.
Such scenarios arise commonly in practice, where simulators often impose hard constraints on parameter values, and a box uniform prior is chosen over an appropriate region. 
Indeed, each task considered in this section uses a box uniform prior on some region, in line with the setup in previous literature, while a MAF with a standard normal base distribution has support everywhere. 
However, this can lead to difficulties, as noted by \citet{greenberg-2019-apt}.
Unlike maximum likelihood training, the classification objective defined by \cref{eq:contrastive-learning-scheme} does not force $ q_{\bfphi} $ to place density in the prior support, but instead dictates only that density ratios are correct in this region; this is an inherent property of training a density estimator in an implicit fashion using a proxy classification task.

The consequence of this detail is that the density estimator can use an arbitrarily small portion of its density to fit the correct ratios if optimization using \cref{eq:contrastive-learning-scheme} means it is advantageous to do so. 
This behaviour occurs commonly in practice, with posterior density `leaking' from the prior support across rounds, such that acceptance rates for rejection sampling using the density estimator can become low.
We stress, however, that this issue only arises with SNPE-C when the posterior support does not match that of the prior.
On the other hand, there is no such similar notion for SRE, since the feed-forward network $ f_{\bfphi} $ defines a density only implicitly and is never evaluated outside of the prior support.   

A full solution to this issue requires either reparameterizing the prior to be unconstrained, or alternatively reparameterizing the density estimator $ q_{\bfphi} $ to be supported only on the same region as the prior. 
We would argue that the former option is heavily influenced by domain-specific factors; it is often natural for practitioners to prescribe hard constraints because simulators fail or run indefinitely for invalid values. 
For example, reparameterizing a strictly positive parameter on a log-scale may be problematic since exponentiated parameter values can lie arbitrarily close to zero, and although technically valid, may be numerically unstable. 
Empirically, we also found that box uniform priors tend to produce sharper posterior estimates than, say, Gaussian priors.
On the other hand, reparameterizing the density estimator to be supported on some fixed range using an invertible squashing function, such as the sigmoid, leads to its own set of numerical issues. 
While theoretically valid, we encountered significant optimization difficulties when testing this approach, with densities often becoming degenerate due to numerical instabilities, and similar observations have been made in the density estimation literature \citep{durkan-2019-cubic}.

We mitigate this issue in practice by fitting SNPE-C in the first round using maximum likelihood, which is possible because we sample parameters from the prior initially. 
This forces the model to place density on the prior support in the first round.
We also found the inductive bias of a MAF beneficial, since a series of elementwise affine transformations makes it difficult to transform a standard normal base distribution in an extreme way. 
In contrast, we found mixture models such as MDNs prone to entire mixture components drifting far from the prior support, while the remaining components are left to model the correct density ratios.

\subsection{Size of the contrasting set}
\vspace{-3pt}
We first examine the role played by the size of the contrasting set. 
To this end, we run both SRE and SNPE-C for $ K \in \curlybr{2, 50, 100} $, and compare metrics across tasks.
Results are shown in \cref{fig:size-of-contrasting-set}.
Across both multiple tasks and multiple metrics, the general trend suggests that increasing the size of the contrasting set results in more efficient inference, if only marginally in some cases.
Computationally, setting $ K = 100 $ means we are effectively evaluating mini-batches of size $ B^{2} $, since the contrasting sets are generated using the full Cartesian product of $ B $ items in the original mini-batch.
However, for $ B = 100 $, we found very little difference in wall-clock time for this operation due to efficient batch computation on the GPU\@.  
These experiments demonstrate the effectiveness of SRE in a relatively low-data regime; \citet{hermans-2019-sre} use one million simulations for a majority of their tasks, whereas the performance of the same algorithm is tested here with orders of magnitude fewer.
Based on these results, the rest of our experiments with SRE and SNPE-C are run using the full mini-batch with $ K = 100 $ to generate the contrasting set. 

\subsection{Comparison of SRE and SNPE-C}
\vspace{-3pt}
The equivalence of SRE and SNPE-C also encourages investigation as to whether reparameterization of the classifier in terms of a conditional density estimator has a noticeable impact on the quality of inference. 
Practitioners may be well-versed in the use of standard classifiers from the modern machine learning toolbox, but less familiar with the recent surge of flexible density estimators based on neural networks. 
Comparison of these approaches may thus guide practical use, since if no discrepancy in performance is observed, SNPE-C has a clear advantage in that a posterior approximation with density evaluation and i.i.d.\ sampling is available at the end of training.
We again refer to \cref{fig:size-of-contrasting-set} for results, which indeed show SNPE-C closely matching SRE\@.
As a control, we also ran SNPE-C with MCMC sampling, and observed essentially the same performance as with i.i.d. sampling.
However, it is important to consider these results in the context of \cref{fig:acceptance-rates-snpe-c}, which visualize rejection sampling acceptance rates across each task for a number of rounds. 
As discussed, SNPE-C leaks density outside of the prior support, with acceptance rates around 1\% meaning i.i.d.\ sampling becomes up to 100 times slower.
Nevertheless, this may still be appreciably faster than MCMC sampling; using a neural density estimator, we may generate large batches consisting of thousands of i.i.d.\ samples in parallel with a few passes of a neural network, while MCMC requires at least one neural network pass to evaluate a transition kernel which may not even result in a single effective sample. 
\looseness=-1

\subsection{Comparison of SRE and SNL}
\vspace{-3pt}
Finally we compare SRE with SNL\@.
Like SRE, SNL is also reliant on MCMC sampling to generate posterior samples in each round.
Results are displayed in \cref{fig:snl-sre-comparison}. 
Overall, SNL seems to outperform SRE, particularly on the M/G/1 task. 
Though we don't do so here, SNL also makes it possible to carry out goodness-of-fit testing to check correctness of its surrogate likelihood model, since sampling observations from the model is straightforward, which is not the case for SRE\@.
However, as mentioned already, if observations are high-dimensional, SRE has a clear advantage in that it need not parameterize a distribution over this high-dimensional space, but only take as input this object to the classifier.

\section{Discussion} 
Recent work has demonstrated cases where likelihood-free inference methods that leverage deep representations can reduce the number of simulations needed for accurate and efficient inference by orders of magnitude.
However, there are several open problems: in their current implementations, the methods described in this paper still (i)~require too many simulations if a simulator takes many days to run, (ii)~have difficulty exploiting multiple observations, and (iii)~don't deal gracefully with failed simulations.
Moreover, the methods in the literature have used a variety of density estimation and density-ratio estimation principles and architectures.
As a result, practitioners facing a new problem domain will have difficulty choosing from the alternatives.

In this work, we identify a strong connection between two strands of the literature, exemplified by SRE and SNPE-C, that were previously compared as unrelated methods. 
Our connection contributes new understanding of these existing approaches, and shows that SRE can and should be generalized to use multi-class classification.

It would be easy to assume that SNPE-C is harder to fit because it uses a conditional density estimator rather than a classifier as in SRE\@. 
However, our control experiments show that\,---\,when trained using the same codebase and classification objective\,---\,the representations give similar results. 
In such cases, using a conditional density estimator should be preferred, because it gives a normalized posterior and removes the need to use MCMC\@. 
However, we also identify open practical problems with SNPE-C for priors with compact support, which mean that MCMC may be necessary in the case of extremely high rejection rates.
We hope that our comparison helps clarify the underlying choices for likelihood-free practitioners, and will motivate researchers to progress the open problems that we have identified.


\section*{Acknowledgements}
The authors would like to thank Theophane Weber for helpful feedback. This work was supported in part by the EPSRC Centre for Doctoral Training in Data Science, funded by the UK Engineering and Physical Sciences Research Council (grant EP/L016427/1) and the University of Edinburgh.

\bibliographystyle{icml2020}
\bibliography{bibliography}

\newpage

\appendix

\onecolumn
\icmltitle{On Contrastive Learning for Likelihood-free Inference \\ Supplementary Material}

\section{Additional experimental results}

\subsection{SRE vs SNPE-C}

\begin{figure*}[!h]
	\centering
    \begin{subfigure}[b]{0.5\textwidth}
        \centering
        \includegraphics[width=0.9\textwidth]{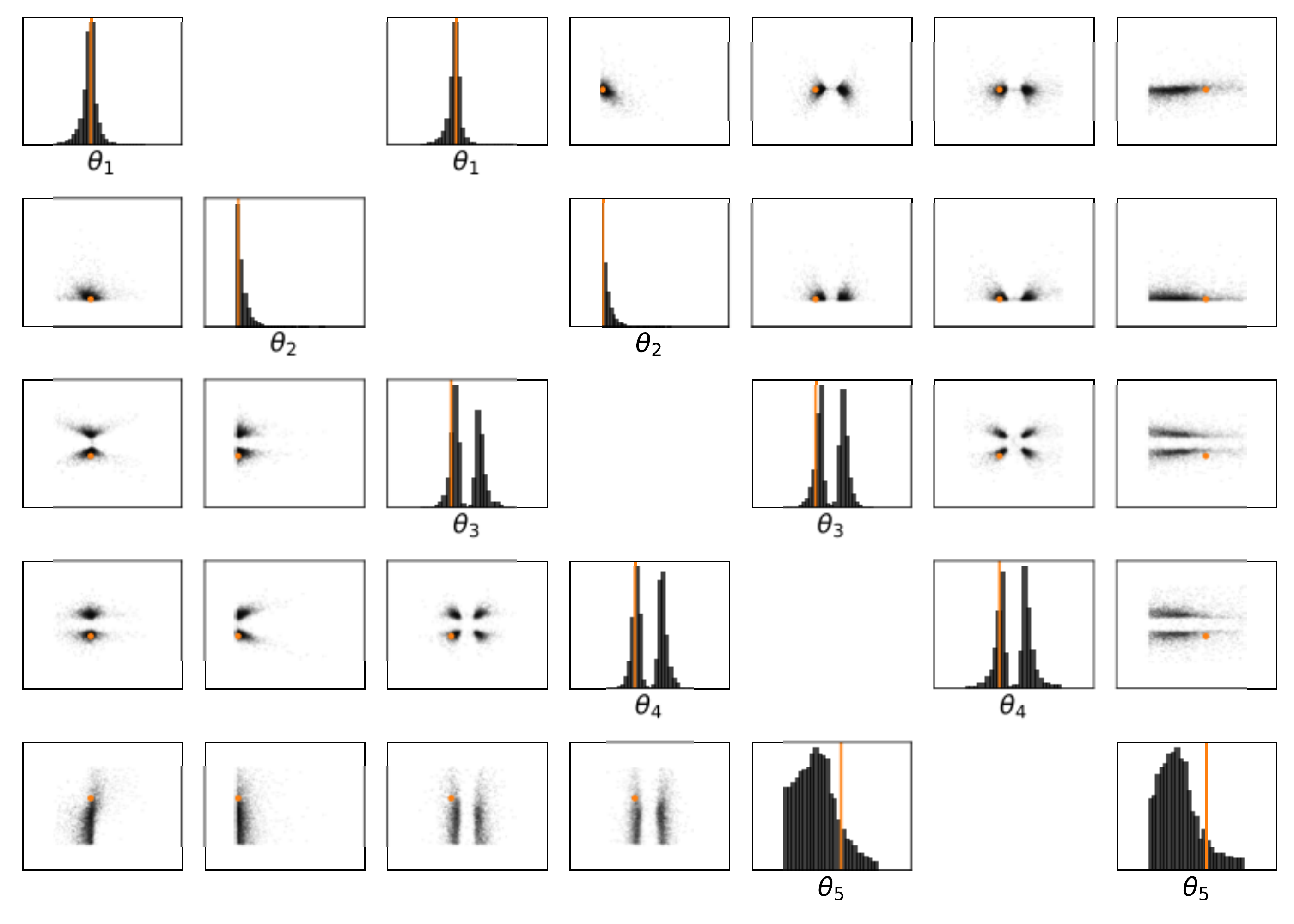}
        \caption{Nonlinear Gaussian}
    \end{subfigure}%
    \begin{subfigure}[b]{0.5\textwidth}
        \centering
        \includegraphics[width=0.9\textwidth]{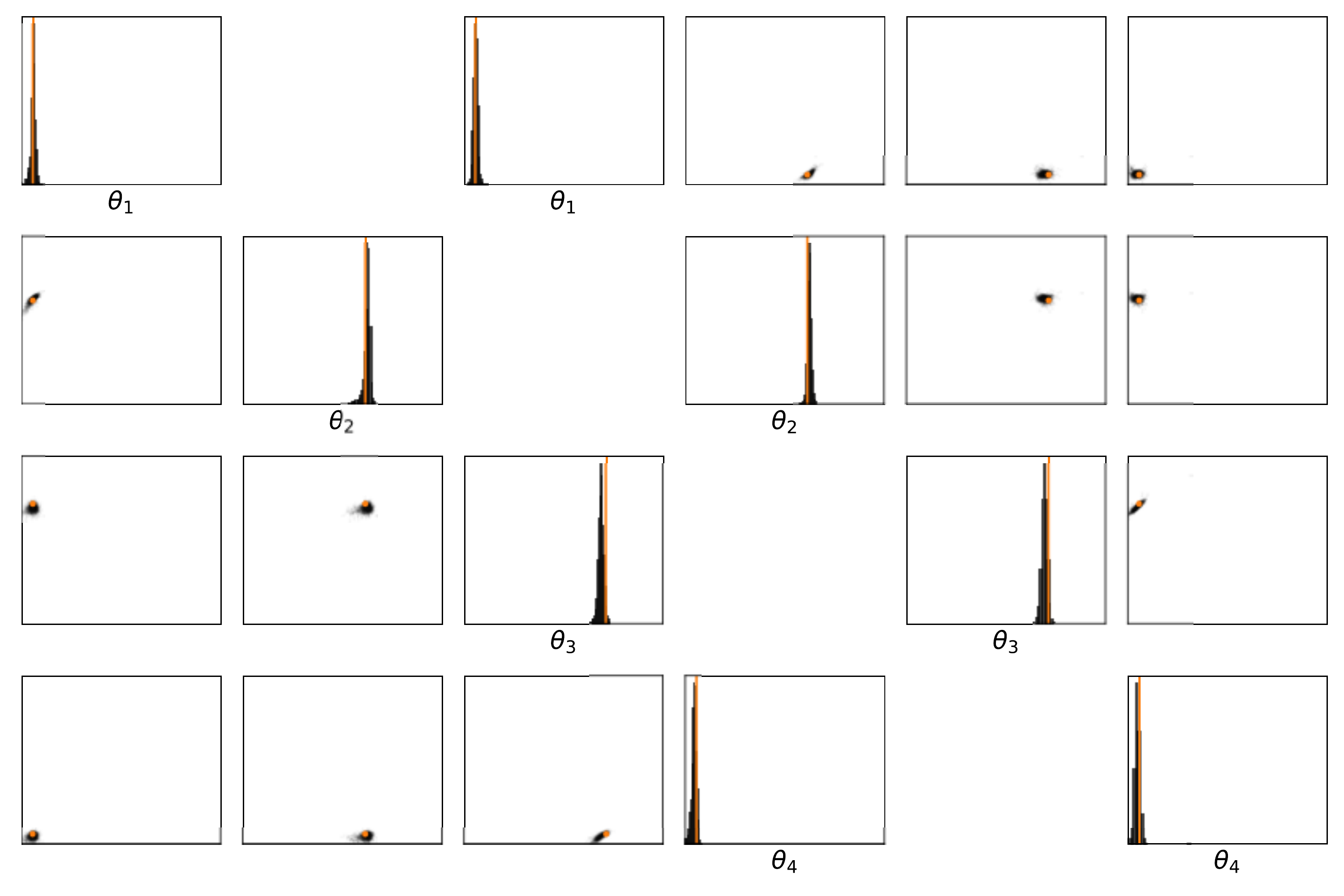}
        \caption{Lotka-Volterra}
    \end{subfigure}
    \begin{subfigure}[b]{0.5\textwidth}
        \centering
        \vspace{17.5pt}
        \includegraphics[width=0.9\textwidth]{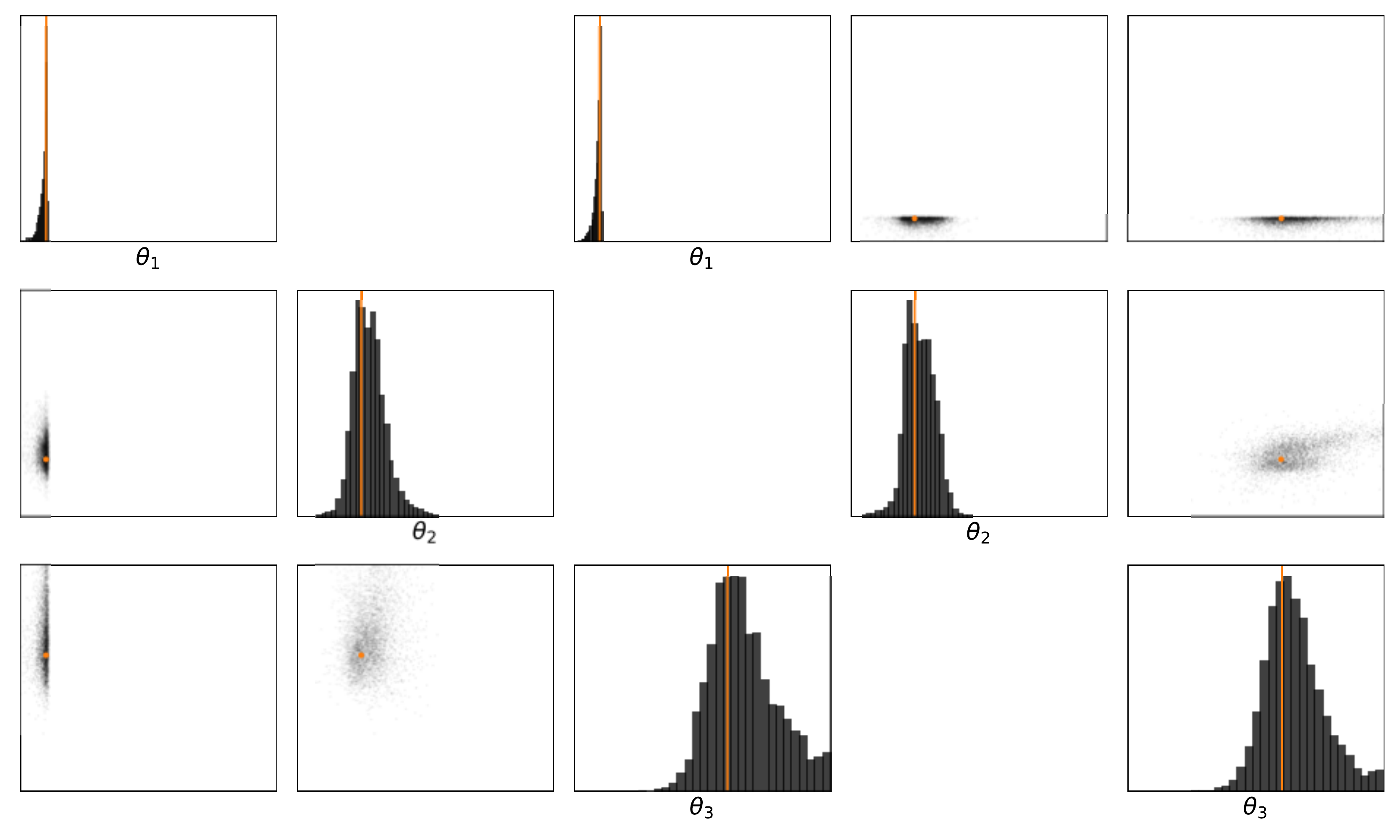}
        \caption{M/G/1}
    \end{subfigure}
	\caption{Comparison of posterior samples for SRE (sub-figure left) and SNPE-C (sub-figure right) on each task. For both methods, we use $ K = 100 $ to generate the contrasting set.}
\end{figure*}

\newpage

\begin{figure*}[!h]
	\centering
    \begin{subfigure}[b]{0.3\textwidth}
        \centering
        \includegraphics[width=0.9\textwidth]{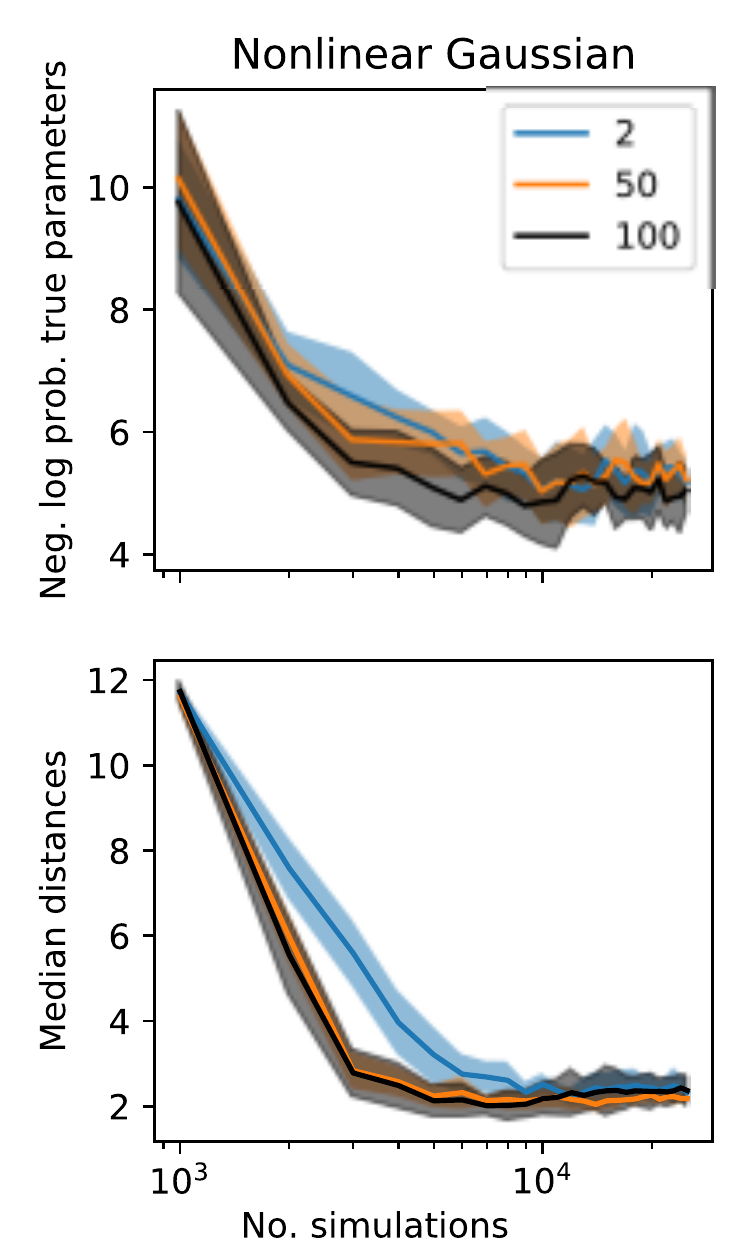}
        \caption{SRE}
    \end{subfigure}%
    \begin{subfigure}[b]{0.3\textwidth}
        \centering
        \includegraphics[width=0.9\textwidth]{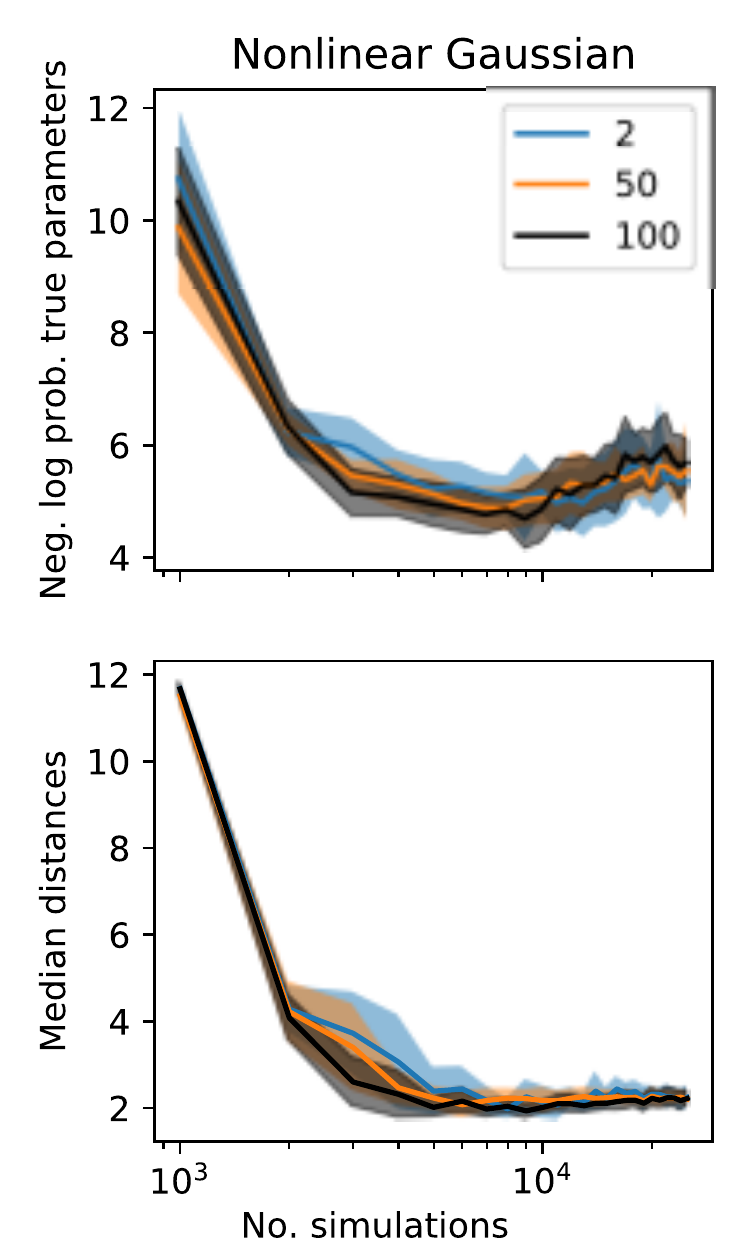}
        \caption{SNPE-C}
    \end{subfigure}
	\caption{Comparison of SRE and SNPE-C metrics on Nonlinear Gaussian task. Before recovering the multimodal posterior, the posteriors can sometimes become too confident in certain parameter settings, leading to the observed negative log likelihood behaviour.}
\end{figure*}

\subsection{SRE vs SNL}

\begin{figure*}[!h]
	\centering
    \begin{subfigure}[b]{0.5\textwidth}
        \centering
        \includegraphics[width=0.9\textwidth]{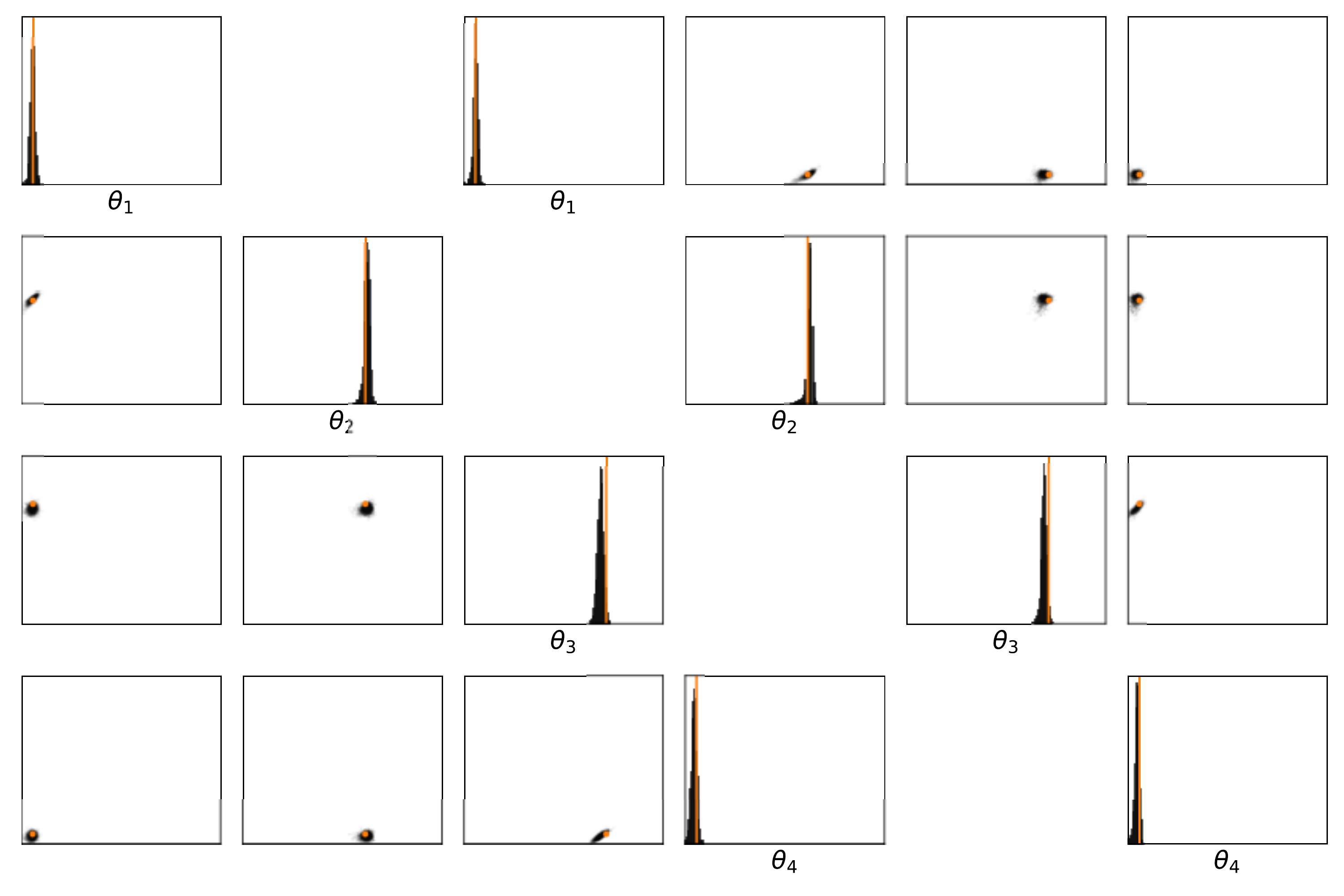}
        \caption{Lotka-Volterra}
    \end{subfigure}%
    \begin{subfigure}[b]{0.5\textwidth}
        \centering
        \includegraphics[width=0.9\textwidth]{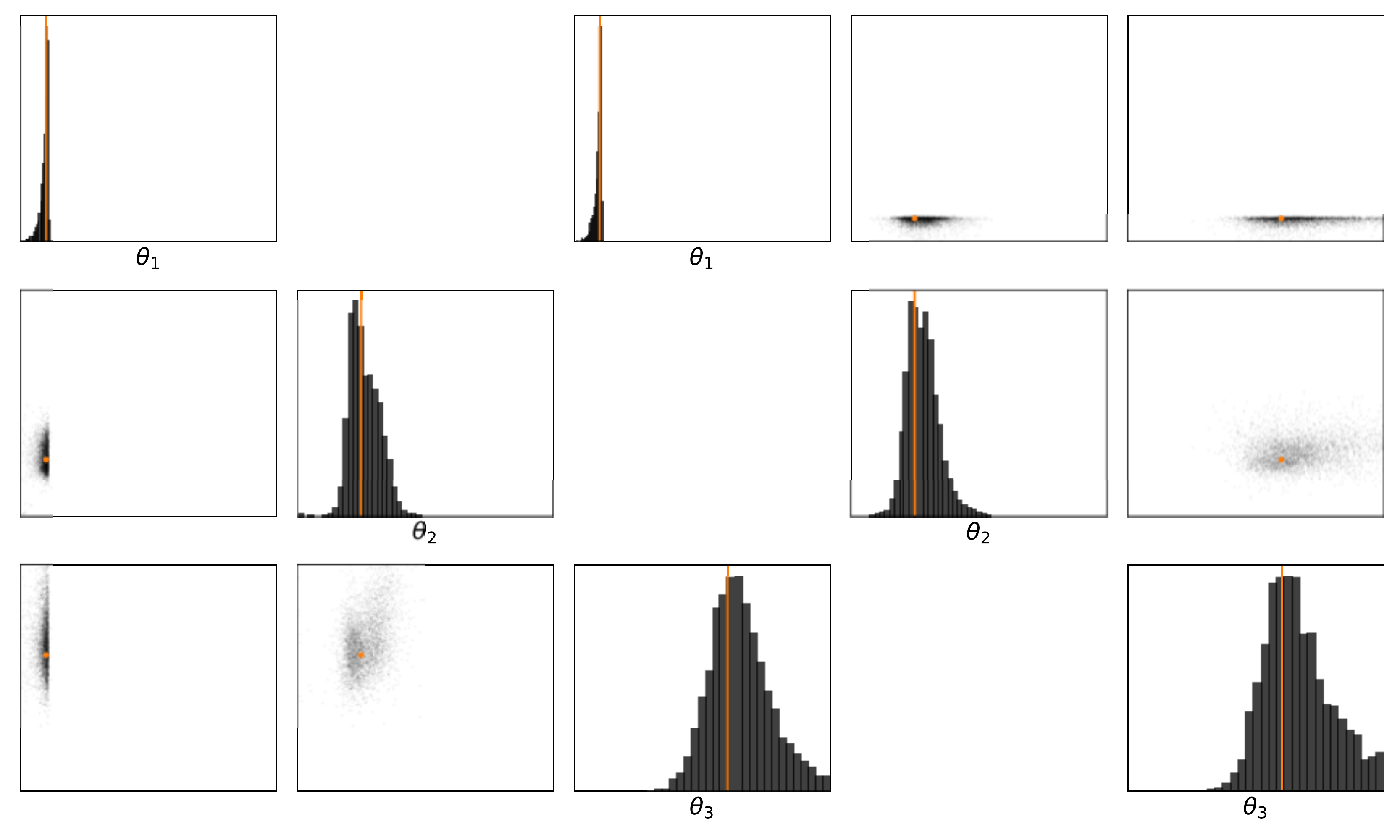}
        \caption{M/G/1}
    \end{subfigure}
	\caption{Comparison of posterior samples for SNL (sub-figure left) and SRE (sub-figure right) on each task.}
\end{figure*}

\newpage

\begin{figure*}[!h]
	\centering
    \includegraphics[width=0.3\textwidth]{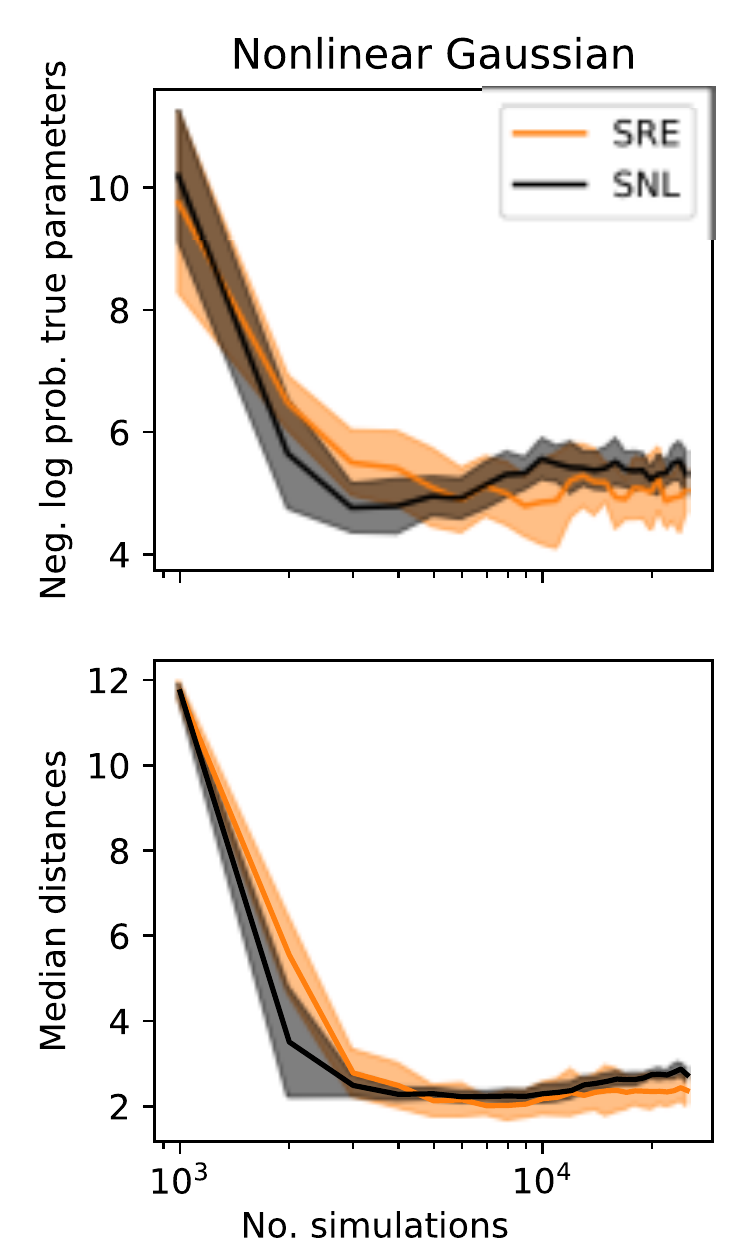}
	\caption{Comparison of Nonlinear Gaussian metrics for SRE and SNL.}
\end{figure*}

\subsection{SNPE-C MCMC}
\begin{figure*}[!h]
	\centering
    \includegraphics[width=0.75\textwidth]{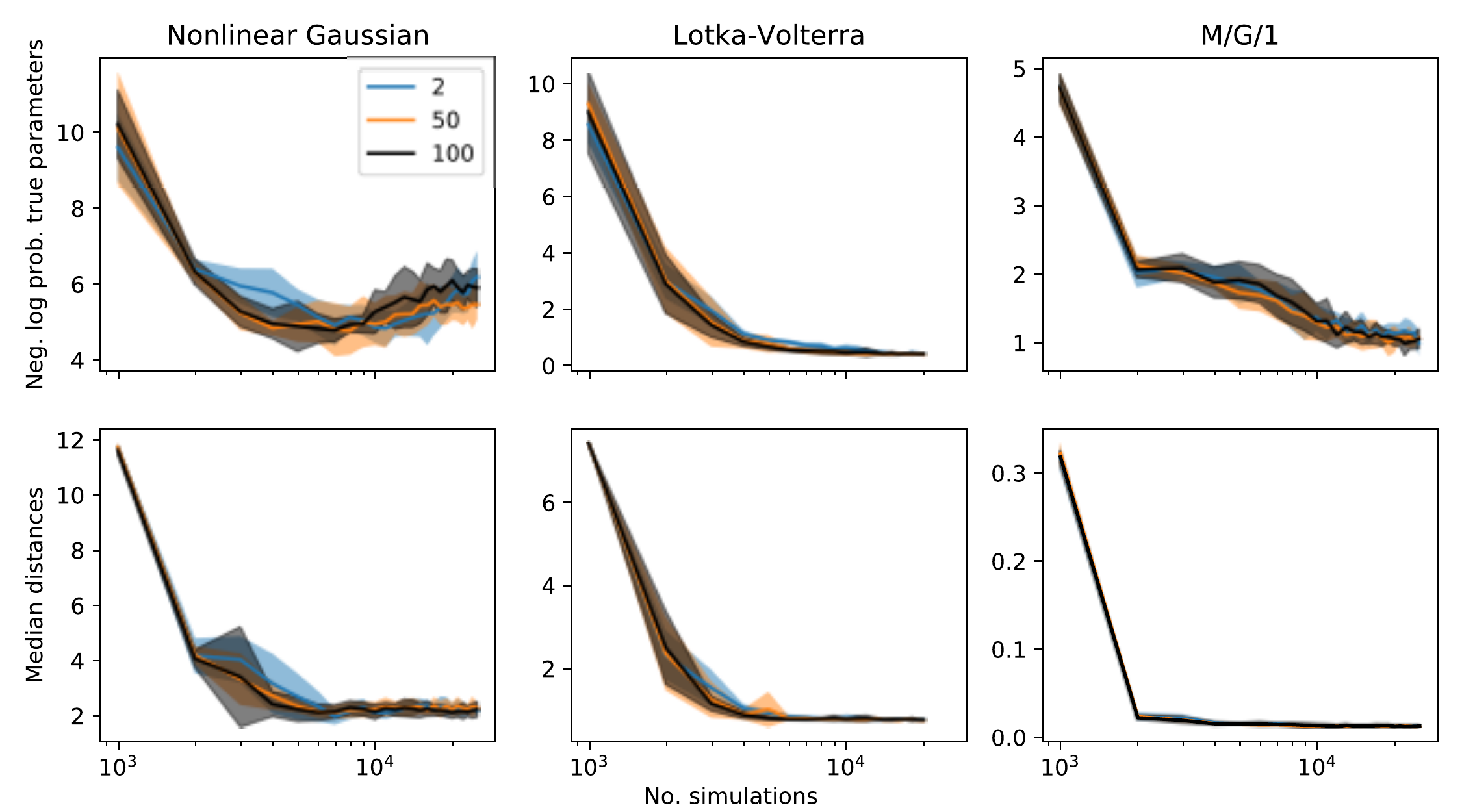}
	\caption{Metrics for Nonlinear Gaussian, Lotka-Volterra, and M/G/1 using SNPE-C with MCMC instead of i.i.d. sampling in each round.}
	\label{fig:snpe-c-mcmc}
\end{figure*}

\newpage

\section{Correctness of SNPE-C for arbitrary proposals $ \tilde{p}(\bftheta) $}
\label{correctness-of-apt}
Assuming optimality of the classifier, we have 
\begin{align}
    f_{\bfphi}(\bfx, \bftheta) &= \log \frac{\tilde{p}(\bftheta \g \bfx)}{\tilde{p}(\bftheta)} + \tilde{c}(\bfx) \\
    &= \log \frac{p(\bftheta \g \bfx)}{p(\bftheta)} + c(\bfx), \text{ where } c(\bfx) = \tilde{c}(\bfx) + \log \frac{p(\bfx)}{\tilde{p}(\bfx)} \label{eq:b}.
\end{align}
Now, since $ f_{\bfphi}(\bfx, \bftheta) = \log \frac{\qphi(\bftheta \g \bfx)}{p(\bftheta)} $, we have
\begin{align}
    \log \frac{\qphi(\bftheta \g \bfx)}{p(\bftheta)} &= \log \frac{p(\bftheta \g \bfx)}{p(\bftheta)} + c(\bfx) \\ 
    \iff \log \qphi(\bftheta \g \bfx) &= \log p(\bftheta \g \bfx) + c(\bfx). 
\end{align}
Exponentiating and then integrating both sides w.r.t.\ $ \bftheta $ gives
\begin{align}
    \int \qphi(\bftheta \g \bfx) \diff \bftheta = \exp\roundbr{c(\bfx)} \int p(\bftheta \g \bfx) \diff \bftheta 
    \,\,
    \Longrightarrow
    \,\,
    c(\bfx) = 0.
\end{align}
Thus for the optimal classifier, we have 
\begin{align}
    \qphi(\bftheta \g \bfx) = p(\bftheta \g \bfx),
\end{align}
and the parameterized conditional density estimator recovers the true posterior for \textit{any} proposal $ \tilde{p}(\bftheta) $.

\end{document}